\documentclass[12pt,a4paper]{article}

\usepackage[margin=1in]{geometry}
\usepackage{amsmath,amssymb,amsthm}
\usepackage{graphicx}
\usepackage{hyperref}
\usepackage{natbib}
\usepackage{booktabs} 
\usepackage{tikz}
\usetikzlibrary{positioning, arrows.meta, calc, shapes.geometric} 
\usepackage{caption}          
\usepackage{algorithm}        
\usepackage{algpseudocode}    
\usepackage{array}
\usepackage{tabularx}  
\usepackage{booktabs}  
\hypersetup{
    hidelinks,
    pdfauthor={Anonymous},
    pdftitle={Your Paper Title},
    pdfsubject={Submission},
    pdfkeywords={keyword1, keyword2}
}

\title{\textbf{PandaAI: A Practical Agent CQ2 for Neuro-symbolic Data Analysis And Integrated Decision-Making in Quantitative Finance}}
\author{\textbf{Yuqi Li} \\
  \textbf{Panda AI} \\
  \texttt{libubai@pandaai.online} \\\and
  \textbf{Siyuan Liu} \\
  \textbf{Panda AI} \\
  \texttt{liusiyuan@pandaai.online} \\\and
  \textbf{Bingjun Liu} \\
  \textbf{Panda AI} \\
  \texttt{liubingjun@pandaai.online} \\}
\date{}

\begin{document}
\maketitle
\begin{abstract}
\noindent While deep learning has excelled in various domains, its application to sequential decision-making in finance remains challenging due to the low Signal-to-Noise Ratio (SNR) and non-stationarity of financial data. Leveraging the reasoning capabilities of Large Language Models (LLMs), we propose \textbf{PandaAI}, a closed-loop neuro-symbolic LLM agent with market regime modeling and constrained alpha generation, which bridges general LLM reasoning with financial rigor and suppresses the financial toxicity of LLM-generated outputs.  To bridge the gap between general linguistic capability and financial rigor, we fine-tune a domain-specific LLM. Furthermore, we integrate this LLM into a modular architecture and form a closed-loop system. Unlike traditional models that optimize isolated prediction metrics, \textbf{PandaAI} is designed as a neuro-symbolic agent that navigates the complex, real-world financial environment with explicit risk awareness. Extensive experiments on CSI 300 stock data show that \textbf{PandaAI} achieves a $18.2\%$ higher Rank IC and $25.7\%$ lower maximum drawdown than state-of-the-art time-series models. Our constrained LLM generation and dual-channel adaptation method provide a general paradigm for LLM deployment in high-stakes sequential decision-making scenarios.
\end{abstract}

\section{Introduction}
Recently, deep learning has obtained great success in many real-world applications, such as facial recognition \cite{Sun_2024_CVPR}, object segmentation \cite{kirillov2023segment}, and natural language processing 
\cite{devlin2019bertpretrainingdeepbidirectional,yenduri2023generativepretrainedtransformercomprehensive}. The financial data brings great challenges to deep learning due to its inherently low Signal-to-Noise Ratio (SNR) and non-stationarity. SNR refers to the relative strength of predictable, economically meaningful patterns (the signal) compared to random, unpredictable fluctuations (the noise) that dominate price movements, and financial price series exhibit strong non-stationarity in the form of trending behavior (near unit-root processes), volatility clustering, structural breaks during economic regime changes, and evolving cross-asset relationships, all of which violate the stationarity assumptions implicit in many standard deep learning architectures.

In this paper, we integrate the quantitative investment method to improve SNR mining the formulaic alpha factor $f$ for decision-making, rather than relying on raw data. The quantitative investment task is modeled as a sequential decision-making process. The goal is to optimize the portfolio weights $\mathbf{w}_t$ to maximize cumulative rewards while satisfying a set of risk constraints $\mathcal{C}$.  A formulaic alpha factor $f$ is a symbolic expression mapping market history to a cross-sectional signal vector $\mathbf{s}_t\in\mathbb{R}^N$ where $N$ represents the number of products in the panel. The search space for $f$ is defined by a context-free grammar involving mathematical operators (e.g., $+$, $-$, $\log$, $\text{rank}$) and market variables. Unlike unconstrained code generation, viable financial factors must adhere to specific structural constraints (e.g., dimensional homogeneity) and risk constraints (e.g., decay rate). We denote the feasible set of factors as $\mathcal{A}_{\text{feasible}} \subset \mathcal{A}_{\text{all}}$.

Financial time series violate the stationary assumption (i.e., joint distributions change over time). We formalize this by introducing a latent regime state $z_t$, which denotes the continuous latent market regime state, which captures the dynamic characteristics of the market (e.g., volatility, liquidity) at time $t$. The market dynamics are governed by a time-varying transition function conditioned on $z_t$. Consequently, a static policy $\pi(a|s)$ inevitably degrades. A market-aware policy must imply $\pi(a|s, z_t)$, dynamically adapting parameters (e.g., risk aversion $\lambda$) based on the inferred regime $z_t$. We summarize the notation in this paper in Table \ref{tab:notations}. In summary, our main contributions are:
\begin{itemize}
    \item \textbf{Constrained MCTS Alpha Mining with LLM Guidance:} We design an LLM-guided constrained MCTS alpha mining framework, integrating financial hard constraints into the full lifecycle of LLM generation to address the financial toxicity issue of factors generated by unconstrained methods.
    \item \textbf{Market Regime Latent Modeling with Dual-Channel Adaptation:} We propose market regime latent variable modeling and a dual-channel adaptation mechanism, compressing high-dimensional market dynamics factors into continuous latent states $z_t$ to achieve unified market perception for LLM's symbolic reasoning and the numerical optimization of quantitative modules.
    
    \item \textbf{Closed-Loop Update System for Quantitative Finance Lifecycle:} We construct a closed-loop update system covering the full lifecycle of quantitative finance, combining fast logical constraint induction and slow parameter adaptation to realize continuous adaptation of the model to non-stationary financial markets, breaking through the limitations of traditional open-loop models.
    
\end{itemize}
\section{Related Work}

\begin{figure*}[t!] 
   \centering
    \includegraphics[width=0.9\textwidth]{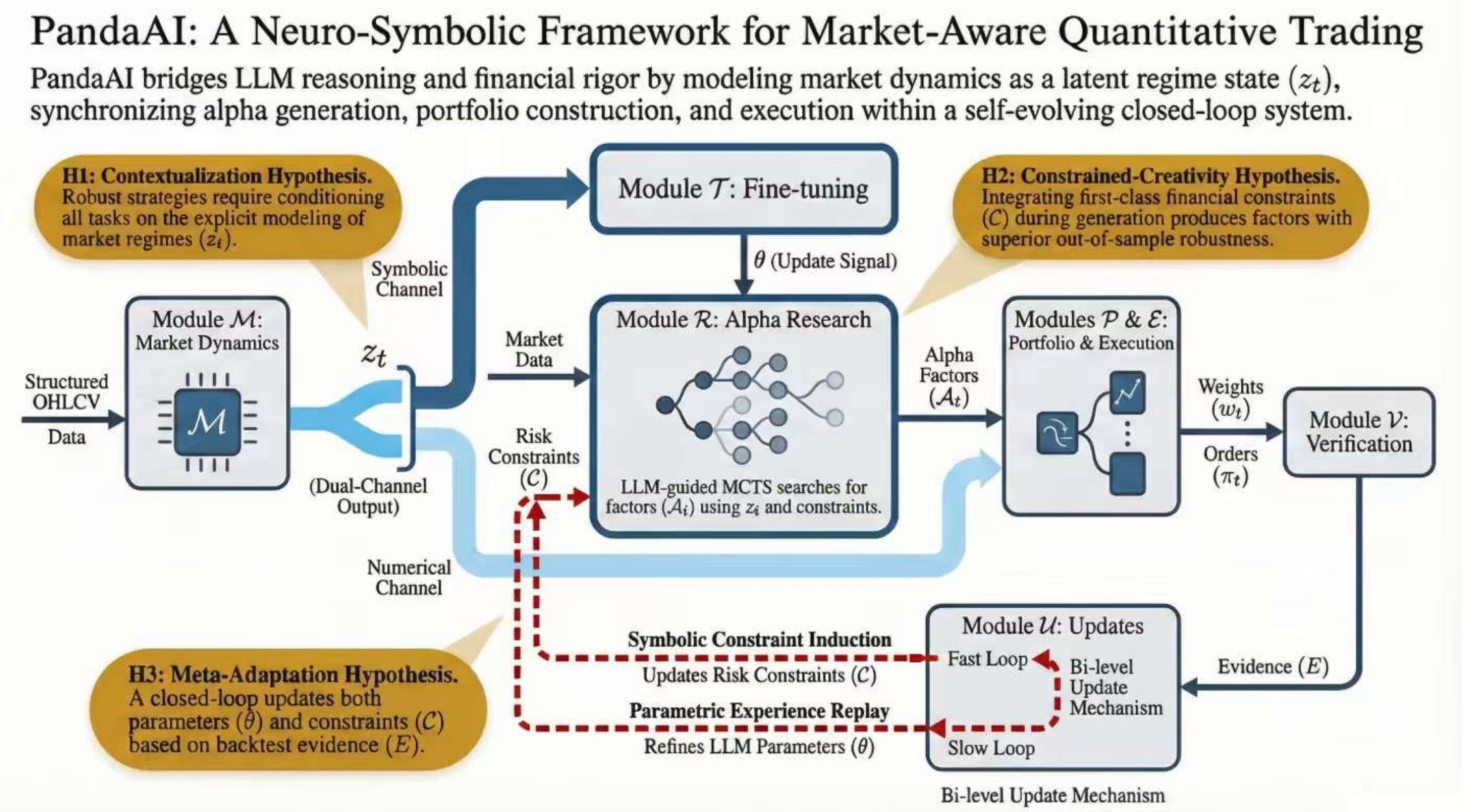} 
    
    \caption{\textbf{Overview of the \textbf{PandaAI} Market-Aware Quantitative Framework.} 
    The system operates as a closed-loop dynamical system spanning six core modules. 
     \textbf{(Left)} The Market Dynamics Module ($\mathcal{M}$) ingests data to generate the regime state $z_t$ (supporting \textbf{H1}). 
     \textbf{(Center)} The Alpha Research Module ($\mathcal{R}$) utilizes LLM-guided MCTS to search for robust factors under constraints $\mathcal{C}$ (supporting \textbf{H2}). 
     \textbf{(Right)} Portfolio ($\mathcal{P}$) and Execution ($\mathcal{E}$) modules actuate decisions conditioned on $z_t$.
     \textbf{(Bottom)} The feedback loop collects Evidence ($E$) via Verification ($\mathcal{V}$) to update parameters $\theta$ and constraints $\mathcal{C}$ (supporting \textbf{H3}).
     \textit{Solid arrows denote Data Flow; Dashed arrows denote Control/Update Flow.}}
     \label{fig:framework_overview}
\end{figure*}
\subsection{Alpha Space Exploration}
Similar to feature engineering in machine learning, automated alpha mining is also a cornerstone of quantitative finance. Before the rise of machine learning, Genetic Programming (GP) \cite{koza1992genetic} performed effectively, although it often inefficiently searched the alpha space. Subsequently, many works have tried their best to cover the full alpha space. \textit{DeepScalper} \cite{sun2022deepscalper} introduces Deep Reinforcement Learning, while \textit{DSO} \cite{petersen2019deep} and \textit{AlphaGen} \cite{yu2023generating} achieve better interpretability using symbolic regression. More recently, since large language models (LLMs) have shown remarkable improvement in their capability for complex semantic understanding and prowess in code-generation, \textit{AlphaGPT} \cite{wang2025alpha} integrates Llama3 70B \cite{grattafiori2024llama3herdmodels} to mine, test, and deploy investment signals (alphas) by translating human intuition into quantitative trading strategies. \cite{shi2025navigating} proposes MCTS-guided exploration to explicitly cover the full alpha space. Despite these significant strides in search capability, robustness remains a primary concern. Generative approaches are susceptible to overfitting, often yielding factors that are mathematically valid but financially toxic (e.g., extreme turnover) due to the absence of continuous, rigorous verification mechanisms.

\subsection{Market Dynamics and Adaptation}
In the broader machine learning community, advanced architectures like \textit{TimesNet} \cite{wu2022timesnet} and \textit{iTransformer} \cite{liu2023itransformer} have set new state-of-the-art standards for handling temporal variations. However, financial markets are inherently non-stationary. The distribution shift derived from market dynamics poses severe challenges to static models \cite{hamilton2020time}. Although \textit{RevIN} \cite{kim2021reversible} and the meta-learning framework \textit{DoubleAdapt}  \cite{zhao2023doubleadapt} have successfully addressed concept drift in stock forecasting, transferring these adaptive mechanisms to LLM-based agents remains underexplored. Contemporary LLM agents typically operate under implicit stationary assumptions, often overlooking the explicit modeling of market dynamics (e.g., regime shifts). This limitation restricts their ability to contextually adapt downstream strategies during periods of market turbulence, such as liquidity crises.




\subsection{Autonomous Agents and Closed-Loop Systems}
The deployment of autonomous agents represents a frontier in AI research. Generalist frameworks \cite{shen2023hugginggpt,park2023generative} demonstrated the immense potential of planning and tool usage. By integrating domain-specific tools, finance-specific agents \cite{li2023tradinggpt,zhang2024multimodal} extended capabilities on finance. Notwithstanding these innovations, current systems predominantly operate in open-loop simulations. They are frequently decoupled from strict financial hard constraints (e.g., leverage limits, transaction costs) and lack systematic feedback loops from execution to model updates. This structural fragmentation limits the potential for synergistic distillation, where insights from one module (e.g., regime detection) could critically inform another (e.g., alpha pruning). To address these limitations, we propose a foundational, market-aware framework that integrates these disparate components into a unified, closed-loop system, enabling holistic optimization across the entire quantitative investment lifecycle.

\section{Methodology}


To bridge the structural fragmentation, we propose a foundational framework and posit that its efficacy stems from three mechanism-driven hypotheses:
\begin{description}
    \item[\textbf{H1 (Contextualization Hypothesis):}] Explicitly modeling market regimes ($z_t$) and conditioning all downstream tasks on it will yield more robust and context-aware strategies than those assuming stationarity.
    \item[\textbf{H2 (Constrained-Creativity Hypothesis):}] Guiding LLM-based alpha generation with first-class financial constraints ($\mathcal{C}$) within an MCTS search will produce factors with superior out-of-sample robustness and lower financial toxicity compared to unconstrained generative methods.
    \item[\textbf{H3 (Meta-Adaptation Hypothesis):}] A closed-loop feedback mechanism that updates both model parameters ($\theta$) and constraint logic ($\mathcal{C}$) based on backtest evidence ($E$) will enable continuous adaptation to non-stationary markets, outperforming static or open-loop systems.
\end{description}
Our framework, as shown in Figure \ref{fig:framework_overview}, is designed to instantiate and test these hypotheses, and Table \ref{tab:hyp} summarizes the relation between our hypotheses and the corresponding mechanisms

\begin{table}[t]
\small
\centering
\begin{tabularx}{\columnwidth}{@{} l l >{\raggedright\arraybackslash}X @{}}
\toprule
\textbf{Hypothesis} & \textbf{Corresponding Modules} & \textbf{Key Implementation Mechanisms} \\
\midrule
\textbf{H1} & $\mathcal{M}, \mathcal{T}, \mathcal{P}, \mathcal{E}$ & 
\textbf{Latent Compression}: Autoencoder for $z_t$ with Dual-Channel Adapters (Soft Tokens \& Scalars) \\
& & \textbf{Prompting}: Continuous soft token injection via Channel 1 \\
& & \textbf{Adaptive Params}: $c(z_t)$ and $\lambda(z_t)$ derived via Channel 2 \\
\midrule
\textbf{H2} & $\mathcal{R}, \mathcal{T}$ & 
\textbf{Pre-Sim Filter}: Hard syntax checking against $G_{\text{forbidden}}$ \\
& & \textbf{Soft Penalty}: Regularization term $-\lambda \cdot \mathbb{I}$ in Eq. (8) \\
& & \textbf{Reasoning Policy}: CoT generation trained via $V_{SFT\_QRA}$ \\
\midrule
\textbf{H3} & $\mathcal{U}, \mathcal{V}, \mathcal{T}$ & 
\textbf{Fast Loop}: LLM-driven logical rule induction for $\mathcal{C}$ \\
& & \textbf{Slow Loop}: Experience Replay with LoRA updates to $\theta$ \\
& & \textbf{Feedback Signal}: Diagnostics from Evidence Bundle $E$ \\
\bottomrule
\end{tabularx}
\caption{Correspondence between Scientific Hypotheses, System Modules, and Implementation Mechanisms.}
\label{tab:hyp}
\end{table}

\subsection{Market Dynamics Module $\mathcal{M}$}
\label{subsec:market_dynamics}

Financial markets are inherently non-stationary, characterized by shifting distributions that render static models obsolete. To address this, we operationalize market awareness not as discrete labels, but as a continuous latent regime manifold. The module $\mathcal{M}$ functions as a compression engine that distills high-dimensional heterogeneous data into a compact, informative state representation $z_t$.

\paragraph{Latent State Construction} 
We utilize \textit{Barra} factors \cite{sheikh1996barra}, which serve as industry-standard risk indicators comprising style (e.g., momentum and volatility) and industry exposures, to characterize market dynamics. We collected these factors over a 10-year horizon. To retain numerical fidelity while reducing noise, a lightweight \textit{Autoencoder} architecture is employed to obtain the low-dimensional $z_t$ that preserves the continuous dynamic properties of the market. This encoder is pre-trained in an unsupervised manner to minimize reconstruction error, ensuring $z_t$ captures the intrinsic manifold of market evolution.

\paragraph{Dual-Channel Adaptation} 
Since $z_t$ must interface with both the symbolic reasoning of the LLM and the numerical optimization of execution modules, we design a Dual-Channel Adapter:\begin{itemize}
    \item \textbf{Symbolic Adapter for LLM (Channel 1):} To make the continuous vector $z_t$ comprehensible to the LLM, we employ a projection MLP that maps $z_t$ into $k$ learnable soft tokens. These tokens are prepended to the LLM's input embedding sequence.
    \item \textbf{Numerical Adapter for Control (Channel 2):} For modules requiring scalar inputs (Portfolio $\mathcal{P}$ and Execution $\mathcal{E}$), a separate feature extraction network maps $z_t$ to specific control parameters (e.g., risk aversion $\lambda_t$, liquidity participation rate $\gamma_t$).
\end{itemize}
This architecture ensures that a unified, consistent market perception $z_t$ drives both the high-level reasoning and low-level control of the agent.

\subsection{LLM-Powered Alpha Research Module $\mathcal{R}$}
\label{subsec:alpha_research}

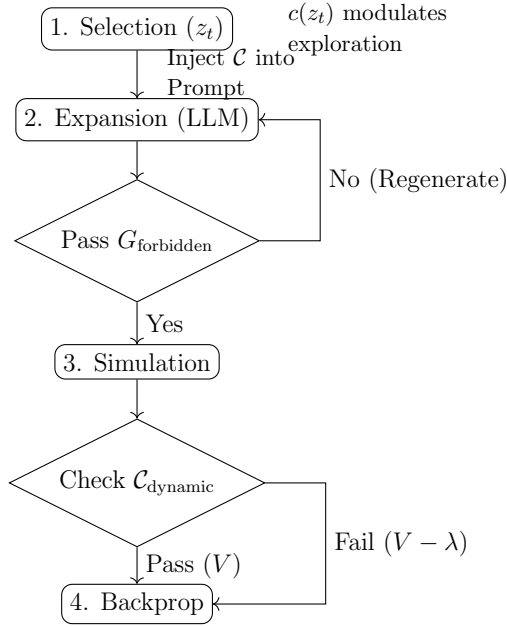
\begin{figure}[t]
\centering
\begin{tikzpicture}[node distance=1.5cm, auto, scale=0.8, every node/.style={transform shape}]
    \node (select) [draw, rectangle, rounded corners] {1. Selection ($z_t$)};
    \node (expand) [draw, rectangle, rounded corners, below of=select] {2. Expansion (LLM)};
    \node (check) [draw, diamond, aspect=2, below of=expand, yshift=-0.5cm] {Pass $G_{\text{forbidden}}$};
    \node (sim) [draw, rectangle, rounded corners, below of=check, yshift=-0.5cm] {3. Simulation};
    \node (eval) [draw, diamond, aspect=2, below of=sim, yshift=-0.5cm] {Check $\mathcal{C}_{\text{dynamic}}$};
    \node (back) [draw, rectangle, rounded corners, below of=eval, yshift=-0.5cm] {4. Backprop};
    
    \draw[->] (select) -- (expand);
    \draw[->] (expand) -- (check);
    \draw[->] (check.south) -- node[right] {Yes} (sim);
    \draw[->] (check.east) -- ++(1,0) |- node[near start, right] {No (Regenerate)} (expand);
    \draw[->] (sim) -- (eval);
    \draw[->] (eval.south) -- node[right] {Pass ($V$)} (back);
    \draw[->] (eval.east) -- ++(1,0) |- node[near start, right] {Fail ($V - \lambda$)} (back);
    
    \node [right of=select, xshift=2.5cm, text width=3cm, align=left] {\small $c(z_t)$ modulates exploration};
    \node [right of=select, yshift=-0.75cm, xshift=0.5cm, text width=3cm, align=left] {\small Inject $\mathcal{C}$ into Prompt};
\end{tikzpicture}
\caption{\textbf{Single MCTS Iteration Flow.} Illustrating where the Constraint Set $\mathcal{C}$ is applied. $G_{\text{forbidden}}$ (a subset of $\mathcal{C}$) acts as a hard filter during Expansion, while dynamic constraints $\mathcal{C}_{\text{dynamic}}$ apply soft penalties during Simulation.}
\label{fig:mcts_flow}
\end{figure}

We conceptualize Alpha Mining not as creative writing, but as a Constrained Search Problem over a Directed Acyclic Graph (DAG) of operators. We implement an LLM-guided Monte Carlo Tree Search (MCTS) framework to navigate this sparse solution space. The process ensures robustness through four quant-specific phases that incorporate the constraint set $\mathcal{C}$ and the market state $z_t$ at distinct checkpoints (visualized in Figure \ref{fig:mcts_flow}); the detailed procedure is presented in Section \ref{sec:cmcts}.

In summary, $\mathcal{R}$ reframes LLM-based financial creativity from an open-ended generation task into a constrained, tree-search-guided reasoning process. This addresses the core limitation of prior generative approaches (Section 3.2): the LLM is not merely a code generator but a reasoning engine whose proposals are continuously subjected to simulation-based financial validation (Eq. 8) within the search loop. The constraint set $\mathcal{C}$ acts primarily as an intrinsic regularizer via prompting and filtering, with residual enforcement via value penalization. This tight integration of symbolic reasoning (LLM), systematic exploration (MCTS), and domain-specific validation (financial backtest) is the key to generating alphas that are both novel and robust. This process directly embodies and operationalizes \textbf{H2}.
\subsection{Fine-Tuning Module $\mathcal{T}$ for \textit{CQ2}} \label{sec:finetuning}

In Module $\mathcal{T}$, the Chain-of-thought Quantitative LLM, named \textit{CQ2}, plays a central role in this framework, handling financial information for textual embedding $h_{txt}$ and generating the formula alpha factor. \textit{CQ2} will be first fine-tuned before other modules.  Unlike generic LLMs, \textit{CQ2} must exhibit two domain-specific capabilities: understanding market dynamics, risk constraints, and alpha semantics, and adapting its reasoning patterns to the market state $z_t$. To achieve this, a two-stage fine-tuning pipeline is designed. \textit{CQ2} utilizes the DeepSeek-Coder-33B architecture \cite{guo2024deepseekcoderlargelanguagemodel}. The fine-tuning procedure consists of Supervised Fine-Tuning (SFT) and Reinforcement Learning from Human Feedback (RLHF). RLHF aligns language models with feedback. Consequently, the fine-tuned LLM can be improved in terms of truthfulness and generalization to user preferences, thereby reducing hallucinations to a great extent, i.e., a $21\%$ vs. $41\%$ hallucination rate before and after, respectively \cite{10.5555/3600270.3602281}. Furthermore, RLHF primarily increases the probability of sampling high-quality rollouts \cite{qi2025evolm}. The detailed fine-tune technique and procedure are demonstrated in Section \ref{sec:apendix_finetune}.

\subsection{Market-Aware Decision Making ($\mathcal{P}$ \& $\mathcal{E}$)}
Instead of static optimization, PandaAI dynamically adjusts the entire trading pipeline based on $z_t$.
\paragraph{Portfolio Optimization ($\mathcal{P}$)} We solve the regime-conditioned convex problem:
\begin{equation}
w_t = \arg\max_{w \in \mathcal{C}_{port}} (w^T s_t - \lambda(z_t) \cdot w^T \Sigma w)
\end{equation}
where $\lambda(z_t)$ automatically increases during high-volatility regimes to prioritize capital preservation.
\paragraph{Execution Control ($\mathcal{E}$)} To minimize implementation shortfall (IS), the execution policy $\pi_{exec}(a|w_t, z_t)$ selects strategies (e.g., TWAP, VWAP) isomorphic to the inferred liquidity profile. This closes the "simulation-to-reality" gap by accounting for market friction.

\subsection{Updates $\mathcal{U}$: Closing the Loop}
\label{subsec:updates}

Operator $\mathcal{U}$ operationalizes \textbf{H3} via a dual-timescale mechanism, enabling PandaAI to adapt to non-stationary environments beyond static pipelines.

\paragraph{Fast Loop (Symbolic Rule Induction)} 
Upon detecting statistically significant failure clusters in $E$, the system triggers symbolic induction. By contrastively prompting the LLM with failed samples ($\text{Sharpe} < 0$) against successful ones within regime $z_t$, $\mathcal{U}$ extracts logical predicates—for instance, identifying that reversal operators underperform during high-momentum phases. These insights are formalized as symbolic rules (e.g., \textit{IF Trend($z_t$) $>$ $\tau$ THEN Ban(Reversal)}) and immediately appended to $\mathcal{C}$.

\paragraph{Slow Loop (Parametric Adaptation)} 
To incorporate long-term market evolution, successful trajectories (verified CoT traces and profitable executions) are stored in an experience replay buffer. We periodically update $\theta$ via LoRA \cite{hu2022lora} with a 5\% data replay ratio \cite{qi2025evolm}. This configuration preserves structural priors while adapting to distribution shifts, effectively mitigating catastrophic forgetting more efficiently than full-parameter fine-tuning. This ensures the model continuously evolves alongside market dynamics without performance degradation.

\section{Experiments}
\label{sec:exp_setting}

\paragraph{Data Partition and Anti-leakage} 
Our data universe consists of the CSI 300 Index constituents. To ensure the integrity of our results against temporal data leakage (a common concern in LLM-based finance), we implement a strict time-series split:
\begin{itemize}
    \item \textbf{Training and SFT Period:} January 1, 2015, to December 31, 2022.
    \item \textbf{Validation/Buffer Period:} January 1, 2023, to December 31, 2023.
    \item \textbf{Out-of-Sample (OOS) Test:} January 1, 2024, to August 31, 2024.
\end{itemize}
Notably, our testing period (2024) is strictly post-release of the DeepSeek-Coder-33B (Nov 2023), ensuring that the agent navigates market dynamics it has never encountered during pre-training. Table \ref{tab:factors} summarizes all the formulas of the alpha factors derived from our approach.

\paragraph{Features and Labels} The input sequence data $x$ has a look-back window of $T=60$ days of OHLCV data. The target label $y_t$ is the cross-sectional standardized 5-day forward return: $y_t = \frac{price_{t+5+1} - price_{t+1}}{price_{t+1}}$. This formulation is consistent with baselines like LSTM \cite{6795963}, Transformer \cite{NIPS2017_3f5ee243}, and StockMixer \cite{10.1609/aaai.v38i8.28681}.

\paragraph{Financial Realism and Metrics} To bridge the ``simulation-to-reality'' gap, our backtest incorporates realistic market frictions: 
\begin{itemize}
    \item \textbf{Transaction Costs:} We apply a double-sided commission of $15$ bps and a slippage of $5$ bps. 
    \item \textbf{Trading Logic:} Rebalancing is performed daily. Daily turnover is capped at $50\%$ through the constraint set $\mathcal{C}$ to suppress ``financially toxic'' high-frequency noise.
\end{itemize}
Performance is evaluated via Information Coefficient (IC), Rank IC, ICIR, Annualized Return (AR), and Maximum Drawdown (MDD). We conduct a 5-group backtest and implement a t-test on the returns. A t-statistic $> 2.0$ ($95\%$ confidence) indicates significant alpha. Formulaic factors used in our experiments are summarized in Table \ref{tab:factors}.

\subsection{Overall Performance}
Table \ref{tab:baselines} compares the performance with the neural network baselines. We observe that pure deep neural networks learn little meaningful pattern from the SNR financial data. The tailored StockMixer can adapt to this data better. It is noted that \textbf{Factor 1} generated by \textbf{PandaAI} outperforms all the neural networks, so our framework can mine the rational formulaic alpha factor from financial data. Furthermore, the t statistic of \textbf{Factor 1} is $9.9667$, so our factor leads to obvious profit.
\begin{table}[t]
\centering
\scriptsize

\begin{tabular}{lccccc}
\toprule
Model     & IC $\uparrow$    & Rank IC $\uparrow$  & ICIR $\uparrow$  & AR  $\uparrow$   & MDD $\uparrow$    \\ 
\midrule
LSTM      & -0.009 & -0.074 & -0.050 & -2.4\% & -62.7\% \\
Transformer   & 0.009  & 0.021  & 0.045  & -0.3\% & -68.9\% \\
StkMixer   & -0.015 & -0.038 & -0.083 & 4.6\%  & -58.8\% \\
\textbf{PandaAI} & \textbf{0.021} & \textbf{0.058} & \textbf{0.193} & \textbf{19.0\%} & \textbf{-44.8\%} \\
\bottomrule
\end{tabular}
\caption{Comparison on CSI 300 with neural network baselines}
\label{tab:baselines}
\end{table}
\subsection{Ablation Study}
In this section, we will explore the effectiveness of each module based on three hypotheses.

\subsubsection{Contextualization Hypothesis Test}
\begin{figure}
    \centering
    \includegraphics[width=0.6\linewidth]{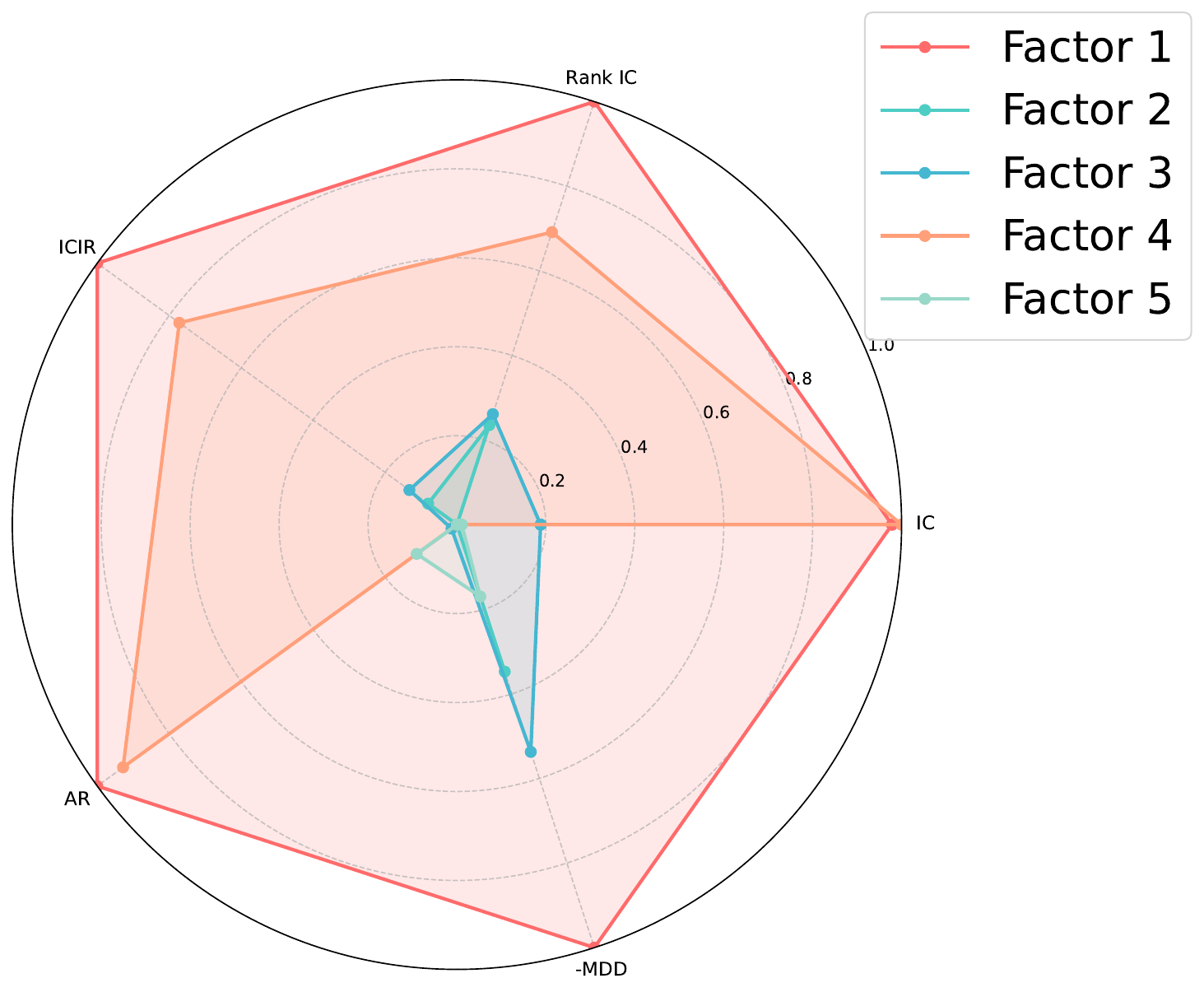}
    \caption{The results of \textbf{Contextualization Hypothesis} from 5 metrics}
    \label{fig:radar}
\end{figure}

To thoroughly investigate the contributions of fine-tuning and the injection of the latent state $  z_t  $, we conduct a series of carefully controlled ablation experiments: \textbf{Factor 2} is generated by the unfine-tuned LLM; \textbf{Factor 3} is generated by the fine-tuned LLM without $  z_t  $; \textbf{Factor 4} is generated for $\mathcal{A}$ without $  z_t  $; \textbf{Factor 5} is generated for $  \mathcal{P}  $ without $z_t$. Figure \ref{fig:radar} visualizes their performance. We can observe that \textbf{Factor 1} from the fully-equipped framework outperforms the factors from ablation studies. This validates the \textbf{H1} is held.

\subsubsection{Constrained-Creativity Hypothesis Test}
\label{subsec:h2_test}

To verify \textbf{H2}, we generated \textbf{Factor 6} by disabling the constraint set $\mathcal{C}$ during the MCTS search. Factor 6 achieved an IC of $0.0207$, Rank IC of $0.0592$, and a raw ICIR of $0.2484$. 

While the unconstrained Factor 6 shows a higher raw ICIR ($0.2484$) than Factor 1 ($0.193$), a deeper financial analysis reveals its \textit{financial toxicity}: its extreme daily turnover ($>80\%$) renders it untradeable in real-world scenarios. Once realistic transaction costs ($15$ bps fee + $5$ bps slippage) are deducted, its net performance drops significantly, failing to maintain consistent profitability. In contrast, \textbf{Factor 1} (generated under constraints $\mathcal{C}$) maintains a healthy balance between predictive power and tradeability. This validates that \textbf{H2} is essential for generating alpha factors that are not just statistically significant but also practically robust and execution-friendly.

\subsubsection{Meta-Adaptation Hypothesis Test}
Meta-Adaptation Hypothesis consists of the fast loop and the slow loop. Firstly, we generate \textbf{Factor 7} without the fast loop. Furthermore, we leave a small-period data for the slow-loop adaptation, and the rest is used for the first fine-tuning. We use this fine-tune LLM to generate \textbf{Factor 8}. Table \ref{tab:meta} shows the result of this ablation study. We conclude that without the fast loop, the quality of mined formulaic alpha decays, and the slow loop make the fine-tune LLM perform similarly with the full data. In a nutshell, the \textbf{H3} is held.

\begin{table}[ht]
\centering

\footnotesize   
\setlength{\tabcolsep}{4pt}  
\begin{tabular}{lccccc}
\toprule
Factor & IC $\uparrow$ & Rank IC $\uparrow$ & ICIR $\uparrow$& AR $\uparrow$ & MDD $\uparrow$\\
\midrule
\textbf{Factor 7} & 0.0133 & 0.5296 & 0.0893 & 16.27\% & -49.88\% \\
\textbf{Factor 8}  & 0.0067 & 0.5236 & 0.0499 & 20.28\% & -44.76\% \\
\bottomrule
\end{tabular}
\caption{Results on Meta-Adaptation Hypothesis Test}
\label{tab:meta}
\end{table}

\section{Conclusion}
In this paper, we presented \textbf{PandaAI}, a neuro-symbolic framework that integrates a fine-tuned domain-specific Large Language Model into a closed-loop system to address the low signal-to-noise ratio and non-stationarity inherent in financial data. By explicitly modeling latent market regimes $  z_t  $ (\textbf{H1}), constraining LLM-guided MCTS search with financial priors to produce robust, low-toxicity alpha factors (\textbf{H2}), and closing the loop via symbolic constraint induction and parametric updates from execution feedback (\textbf{H3}), \textbf{PandaAI} enables adaptive, context-aware quantitative decision-making across market dynamics, alpha mining, portfolio optimization, and realistic execution. 

\newpage
\bibliographystyle{plain}
\bibliography{custom} 

@article{koza1992genetic,
  title={Genetic programming: on the programming of computers by means of natural selection Cambridge},
  author={Koza, John R},
  journal={MA: MIT Press.[Google Scholar]},
  year={1992}
}

@inproceedings{sun2022deepscalper,
  title={DeepScalper: A risk-aware reinforcement learning framework to capture fleeting intraday trading opportunities},
  author={Sun, Shuo and Xue, Wanqi and Wang, Rundong and He, Xu and Zhu, Junlei and Li, Jian and An, Bo},
  booktitle={Proceedings of the 31st ACM International Conference on Information \& Knowledge Management},
  pages={1858--1867},
  year={2022}
}

@article{petersen2019deep,
  title={Deep symbolic regression: Recovering mathematical expressions from data via risk-seeking policy gradients},
  author={Petersen, Brenden K and Landajuela, Mikel and Mundhenk, T Nathan and Santiago, Claudio P and Kim, Soo K and Kim, Joanne T},
  journal={arXiv preprint arXiv:1912.04871},
  year={2019}
}

@inproceedings{yu2023generating,
  title={Generating synergistic formulaic alpha collections via reinforcement learning},
  author={Yu, Shuo and Xue, Hongyan and Ao, Xiang and Pan, Feiyang and He, Jia and Tu, Dandan and He, Qing},
  booktitle={Proceedings of the 29th ACM SIGKDD Conference on Knowledge Discovery and Data Mining},
  pages={5476--5486},
  year={2023}
}

@inproceedings{wang2025alpha,
  title={Alpha-gpt: Human-ai interactive alpha mining for quantitative investment},
  author={Wang, Saizhuo and Yuan, Hang and Zhou, Leon and Ni, Lionel and Shum, Heung Yeung and Guo, Jian},
  booktitle={Proceedings of the 2025 Conference on Empirical Methods in Natural Language Processing: System Demonstrations},
  pages={196--206},
  year={2025}
}

@article{shi2025navigating,
  title={Navigating the Alpha Jungle: An LLM-Powered MCTS Framework for Formulaic Factor Mining},
  author={Shi, Yu and Duan, Yitong and Li, Jian},
  journal={arXiv preprint arXiv:2505.11122},
  year={2025}
}

@book{hamilton2020time,
  title={Time series analysis},
  author={Hamilton, James D},
  year={2020},
  publisher={Princeton university press}
}

@article{wu2022timesnet,
  title={Timesnet: Temporal 2d-variation modeling for general time series analysis},
  author={Wu, Haixu and Hu, Tengge and Liu, Yong and Zhou, Hang and Wang, Jianmin and Long, Mingsheng},
  journal={arXiv preprint arXiv:2210.02186},
  year={2022}
}

@article{liu2023itransformer,
  title={itransformer: Inverted transformers are effective for time series forecasting},
  author={Liu, Yong and Hu, Tengge and Zhang, Haoran and Wu, Haixu and Wang, Shiyu and Ma, Lintao and Long, Mingsheng},
  journal={arXiv preprint arXiv:2310.06625},
  year={2023}
}

@inproceedings{kim2021reversible,
  title={Reversible instance normalization for accurate time-series forecasting against distribution shift},
  author={Kim, Taesung and Kim, Jinhee and Tae, Yunwon and Park, Cheonbok and Choi, Jang-Ho and Choo, Jaegul},
  booktitle={International conference on learning representations},
  year={2021}
}

@inproceedings{zhao2023doubleadapt,
  title={Doubleadapt: A meta-learning approach to incremental learning for stock trend forecasting},
  author={Zhao, Lifan and Kong, Shuming and Shen, Yanyan},
  booktitle={Proceedings of the 29th ACM SIGKDD Conference on Knowledge Discovery and Data Mining},
  pages={3492--3503},
  year={2023}
}

@inproceedings{10.5555/3495724.3495977,
author = {Stiennon, Nisan and Ouyang, Long and Wu, Jeff and Ziegler, Daniel M. and Lowe, Ryan and Voss, Chelsea and Radford, Alec and Amodei, Dario and Christiano, Paul},
title = {Learning to summarize from human feedback},
year = {2020},
isbn = {9781713829546},
publisher = {Curran Associates Inc.},
address = {Red Hook, NY, USA},
booktitle = {Proceedings of the 34th International Conference on Neural Information Processing Systems},
articleno = {253},
numpages = {14},
location = {Vancouver, BC, Canada},
series = {NIPS '20}
}

@article{shen2023hugginggpt,
  title={Hugginggpt: Solving ai tasks with chatgpt and its friends in hugging face},
  author={Shen, Yongliang and Song, Kaitao and Tan, Xu and Li, Dongsheng and Lu, Weiming and Zhuang, Yueting},
  journal={Advances in Neural Information Processing Systems},
  volume={36},
  pages={38154--38180},
  year={2023}
}

@inproceedings{park2023generative,
  title={Generative agents: Interactive simulacra of human behavior},
  author={Park, Joon Sung and O'Brien, Joseph and Cai, Carrie Jun and Morris, Meredith Ringel and Liang, Percy and Bernstein, Michael S},
  booktitle={Proceedings of the 36th annual acm symposium on user interface software and technology},
  pages={1--22},
  year={2023}
}

@article{li2023tradinggpt,
  title={Tradinggpt: Multi-agent system with layered memory and distinct characters for enhanced financial trading performance},
  author={Li, Yang and Yu, Yangyang and Li, Haohang and Chen, Zhi and Khashanah, Khaldoun},
  journal={arXiv preprint arXiv:2309.03736},
  year={2023}
}

@inproceedings{zhang2024multimodal,
  title={A multimodal foundation agent for financial trading: Tool-augmented, diversified, and generalist},
  author={Zhang, Wentao and Zhao, Lingxuan and Xia, Haochong and Sun, Shuo and Sun, Jiaze and Qin, Molei and Li, Xinyi and Zhao, Yuqing and Zhao, Yilei and Cai, Xinyu and others},
  booktitle={Proceedings of the 30th acm sigkdd conference on knowledge discovery and data mining},
  pages={4314--4325},
  year={2024}
}

@inproceedings{
qi2025evolm,
title={Evo{LM}: In Search of Lost Language Model Training Dynamics},
author={Zhenting Qi and Fan Nie and Alexandre Alahi and James Zou and Himabindu Lakkaraju and Yilun Du and Eric P. Xing and Sham M. Kakade and Hanlin Zhang},
booktitle={The Thirty-ninth Annual Conference on Neural Information Processing Systems},
year={2025},
url={https://openreview.net/forum?id=B6bE2GC71a}
}

@inproceedings{
hu2022lora,
title={Lo{RA}: Low-Rank Adaptation of Large Language Models},
author={Edward J Hu and yelong shen and Phillip Wallis and Zeyuan Allen-Zhu and Yuanzhi Li and Shean Wang and Lu Wang and Weizhu Chen},
booktitle={International Conference on Learning Representations},
year={2022},
url={https://openreview.net/forum?id=nZeVKeeFYf9}
}

@inproceedings{li-etal-2024-revisiting,
    title = "Revisiting Catastrophic Forgetting in Large Language Model Tuning",
    author = "Li, Hongyu  and
      Ding, Liang  and
      Fang, Meng  and
      Tao, Dacheng",
    editor = "Al-Onaizan, Yaser  and
      Bansal, Mohit  and
      Chen, Yun-Nung",
    booktitle = "Findings of the Association for Computational Linguistics: EMNLP 2024",
    month = nov,
    year = "2024",
    address = "Miami, Florida, USA",
    publisher = "Association for Computational Linguistics",
    url = "https://aclanthology.org/2024.findings-emnlp.249/",
    doi = "10.18653/v1/2024.findings-emnlp.249",
    pages = "4297--4308"
}

@inproceedings{10.5555/3600270.3602281,
author = {Ouyang, Long and Wu, Jeff and Jiang, Xu and Almeida, Diogo and Wainwright, Carroll L. and Mishkin, Pamela and Zhang, Chong and Agarwal, Sandhini and Slama, Katarina and Ray, Alex and Schulman, John and Hilton, Jacob and Kelton, Fraser and Miller, Luke and Simens, Maddie and Askell, Amanda and Welinder, Peter and Christiano, Paul and Leike, Jan and Lowe, Ryan},
title = {Training language models to follow instructions with human feedback},
year = {2022},
isbn = {9781713871088},
publisher = {Curran Associates Inc.},
address = {Red Hook, NY, USA},
booktitle = {Proceedings of the 36th International Conference on Neural Information Processing Systems},
articleno = {2011},
numpages = {15},
location = {New Orleans, LA, USA},
series = {NIPS '22}
}

@misc{schulman2017proximalpolicyoptimizationalgorithms,
      title={Proximal Policy Optimization Algorithms}, 
      author={John Schulman and Filip Wolski and Prafulla Dhariwal and Alec Radford and Oleg Klimov},
      year={2017},
      eprint={1707.06347},
      archivePrefix={arXiv},
      primaryClass={cs.LG},
      url={https://arxiv.org/abs/1707.06347}, 
}

@misc{deepseekai2025deepseekr1incentivizingreasoningcapability,
      title={DeepSeek-R1: Incentivizing Reasoning Capability in LLMs via Reinforcement Learning}, 
      author={DeepSeek-AI},
      year={2025},
      eprint={2501.12948},
      archivePrefix={arXiv},
      primaryClass={cs.CL},
      url={https://arxiv.org/abs/2501.12948}, 
}

@misc{guo2024deepseekcoderlargelanguagemodel,
      title={DeepSeek-Coder: When the Large Language Model Meets Programming -- The Rise of Code Intelligence}, 
      author={Daya Guo and Qihao Zhu and Dejian Yang and Zhenda Xie and Kai Dong and Wentao Zhang and Guanting Chen and Xiao Bi and Y. Wu and Y. K. Li and Fuli Luo and Yingfei Xiong and Wenfeng Liang},
      year={2024},
      eprint={2401.14196},
      archivePrefix={arXiv},
      primaryClass={cs.SE},
      url={https://arxiv.org/abs/2401.14196}, 
}

@misc{grattafiori2024llama3herdmodels,
      title={The Llama 3 Herd of Models}, 
      author={Aaron Grattafiori and Abhimanyu Dubey and Abhinav Jauhri and Abhinav Pandey and Abhishek Kadian and Ahmad Al-Dahle and Aiesha Letman and Akhil Mathur and Alan Schelten and Alex Vaughan and Amy Yang and Angela Fan and Anirudh Goyal and Anthony Hartshorn and Aobo Yang and Archi Mitra and Archie Sravankumar and Artem Korenev and Arthur Hinsvark and Arun Rao and Aston Zhang and Aurelien Rodriguez and Austen Gregerson and Ava Spataru and Baptiste Roziere and Bethany Biron and Binh Tang and Bobbie Chern and Charlotte Caucheteux and Chaya Nayak and Chloe Bi and Chris Marra and Chris McConnell and Christian Keller and Christophe Touret and Chunyang Wu and Corinne Wong and Cristian Canton Ferrer and Cyrus Nikolaidis and Damien Allonsius and Daniel Song and Danielle Pintz and Danny Livshits and Danny Wyatt and David Esiobu and Dhruv Choudhary and Dhruv Mahajan and Diego Garcia-Olano and Diego Perino and Dieuwke Hupkes and Egor Lakomkin and Ehab AlBadawy and Elina Lobanova and Emily Dinan and Eric Michael Smith and Filip Radenovic and Francisco Guzmán and Frank Zhang and Gabriel Synnaeve and Gabrielle Lee and Georgia Lewis Anderson and Govind Thattai and Graeme Nail and Gregoire Mialon and Guan Pang and Guillem Cucurell and Hailey Nguyen and Hannah Korevaar and Hu Xu and Hugo Touvron and Iliyan Zarov and Imanol Arrieta Ibarra and Isabel Kloumann and Ishan Misra and Ivan Evtimov and Jack Zhang and Jade Copet and Jaewon Lee and Jan Geffert and Jana Vranes and Jason Park and Jay Mahadeokar and Jeet Shah and Jelmer van der Linde and Jennifer Billock and Jenny Hong and Jenya Lee and Jeremy Fu and Jianfeng Chi and Jianyu Huang and Jiawen Liu and Jie Wang and Jiecao Yu and Joanna Bitton and Joe Spisak and Jongsoo Park and Joseph Rocca and Joshua Johnstun and Joshua Saxe and Junteng Jia and Kalyan Vasuden Alwala and Karthik Prasad and Kartikeya Upasani and Kate Plawiak and Ke Li and Kenneth Heafield and Kevin Stone and Khalid El-Arini and Krithika Iyer and Kshitiz Malik and Kuenley Chiu and Kunal Bhalla and Kushal Lakhotia and Lauren Rantala-Yeary and Laurens van der Maaten and Lawrence Chen and Liang Tan and Liz Jenkins and Louis Martin and Lovish Madaan and Lubo Malo and Lukas Blecher and Lukas Landzaat and Luke de Oliveira and Madeline Muzzi and Mahesh Pasupuleti and Mannat Singh and Manohar Paluri and Marcin Kardas and Maria Tsimpoukelli and Mathew Oldham and Mathieu Rita and Maya Pavlova and Melanie Kambadur and Mike Lewis and Min Si and Mitesh Kumar Singh and Mona Hassan and Naman Goyal and Narjes Torabi and Nikolay Bashlykov and Nikolay Bogoychev and Niladri Chatterji and Ning Zhang and Olivier Duchenne and Onur Çelebi and Patrick Alrassy and Pengchuan Zhang and Pengwei Li and Petar Vasic and Peter Weng and Prajjwal Bhargava and Pratik Dubal and Praveen Krishnan and Punit Singh Koura and Puxin Xu and Qing He and Qingxiao Dong and Ragavan Srinivasan and Raj Ganapathy and Ramon Calderer and Ricardo Silveira Cabral and Robert Stojnic and Roberta Raileanu and Rohan Maheswari and Rohit Girdhar and Rohit Patel and Romain Sauvestre and Ronnie Polidoro and Roshan Sumbaly and Ross Taylor and Ruan Silva and Rui Hou and Rui Wang and Saghar Hosseini and Sahana Chennabasappa and Sanjay Singh and Sean Bell and Seohyun Sonia Kim and Sergey Edunov and Shaoliang Nie and Sharan Narang and Sharath Raparthy and Sheng Shen and Shengye Wan and Shruti Bhosale and Shun Zhang and Simon Vandenhende and Soumya Batra and Spencer Whitman and Sten Sootla and Stephane Collot and Suchin Gururangan and Sydney Borodinsky and Tamar Herman and Tara Fowler and Tarek Sheasha and Thomas Georgiou and Thomas Scialom and Tobias Speckbacher and Todor Mihaylov and Tong Xiao and Ujjwal Karn and Vedanuj Goswami and Vibhor Gupta and Vignesh Ramanathan and Viktor Kerkez and Vincent Gonguet and Virginie Do and Vish Vogeti and Vítor Albiero and Vladan Petrovic and Weiwei Chu and Wenhan Xiong and Wenyin Fu and Whitney Meers and Xavier Martinet and Xiaodong Wang and Xiaofang Wang and Xiaoqing Ellen Tan and Xide Xia and Xinfeng Xie and Xuchao Jia and Xuewei Wang and Yaelle Goldschlag and Yashesh Gaur and Yasmine Babaei and Yi Wen and Yiwen Song and Yuchen Zhang and Yue Li and Yuning Mao and Zacharie Delpierre Coudert and Zheng Yan and Zhengxing Chen and Zoe Papakipos and Aaditya Singh and Aayushi Srivastava and Abha Jain and Adam Kelsey and Adam Shajnfeld and Adithya Gangidi and Adolfo Victoria and Ahuva Goldstand and Ajay Menon and Ajay Sharma and Alex Boesenberg and Alexei Baevski and Allie Feinstein and Amanda Kallet and Amit Sangani and Amos Teo and Anam Yunus and Andrei Lupu and Andres Alvarado and Andrew Caples and Andrew Gu and Andrew Ho and Andrew Poulton and Andrew Ryan and Ankit Ramchandani and Annie Dong and Annie Franco and Anuj Goyal and Aparajita Saraf and Arkabandhu Chowdhury and Ashley Gabriel and Ashwin Bharambe and Assaf Eisenman and Azadeh Yazdan and Beau James and Ben Maurer and Benjamin Leonhardi and Bernie Huang and Beth Loyd and Beto De Paola and Bhargavi Paranjape and Bing Liu and Bo Wu and Boyu Ni and Braden Hancock and Bram Wasti and Brandon Spence and Brani Stojkovic and Brian Gamido and Britt Montalvo and Carl Parker and Carly Burton and Catalina Mejia and Ce Liu and Changhan Wang and Changkyu Kim and Chao Zhou and Chester Hu and Ching-Hsiang Chu and Chris Cai and Chris Tindal and Christoph Feichtenhofer and Cynthia Gao and Damon Civin and Dana Beaty and Daniel Kreymer and Daniel Li and David Adkins and David Xu and Davide Testuggine and Delia David and Devi Parikh and Diana Liskovich and Didem Foss and Dingkang Wang and Duc Le and Dustin Holland and Edward Dowling and Eissa Jamil and Elaine Montgomery and Eleonora Presani and Emily Hahn and Emily Wood and Eric-Tuan Le and Erik Brinkman and Esteban Arcaute and Evan Dunbar and Evan Smothers and Fei Sun and Felix Kreuk and Feng Tian and Filippos Kokkinos and Firat Ozgenel and Francesco Caggioni and Frank Kanayet and Frank Seide and Gabriela Medina Florez and Gabriella Schwarz and Gada Badeer and Georgia Swee and Gil Halpern and Grant Herman and Grigory Sizov and Guangyi and Zhang and Guna Lakshminarayanan and Hakan Inan and Hamid Shojanazeri and Han Zou and Hannah Wang and Hanwen Zha and Haroun Habeeb and Harrison Rudolph and Helen Suk and Henry Aspegren and Hunter Goldman and Hongyuan Zhan and Ibrahim Damlaj and Igor Molybog and Igor Tufanov and Ilias Leontiadis and Irina-Elena Veliche and Itai Gat and Jake Weissman and James Geboski and James Kohli and Janice Lam and Japhet Asher and Jean-Baptiste Gaya and Jeff Marcus and Jeff Tang and Jennifer Chan and Jenny Zhen and Jeremy Reizenstein and Jeremy Teboul and Jessica Zhong and Jian Jin and Jingyi Yang and Joe Cummings and Jon Carvill and Jon Shepard and Jonathan McPhie and Jonathan Torres and Josh Ginsburg and Junjie Wang and Kai Wu and Kam Hou U and Karan Saxena and Kartikay Khandelwal and Katayoun Zand and Kathy Matosich and Kaushik Veeraraghavan and Kelly Michelena and Keqian Li and Kiran Jagadeesh and Kun Huang and Kunal Chawla and Kyle Huang and Lailin Chen and Lakshya Garg and Lavender A and Leandro Silva and Lee Bell and Lei Zhang and Liangpeng Guo and Licheng Yu and Liron Moshkovich and Luca Wehrstedt and Madian Khabsa and Manav Avalani and Manish Bhatt and Martynas Mankus and Matan Hasson and Matthew Lennie and Matthias Reso and Maxim Groshev and Maxim Naumov and Maya Lathi and Meghan Keneally and Miao Liu and Michael L. Seltzer and Michal Valko and Michelle Restrepo and Mihir Patel and Mik Vyatskov and Mikayel Samvelyan and Mike Clark and Mike Macey and Mike Wang and Miquel Jubert Hermoso and Mo Metanat and Mohammad Rastegari and Munish Bansal and Nandhini Santhanam and Natascha Parks and Natasha White and Navyata Bawa and Nayan Singhal and Nick Egebo and Nicolas Usunier and Nikhil Mehta and Nikolay Pavlovich Laptev and Ning Dong and Norman Cheng and Oleg Chernoguz and Olivia Hart and Omkar Salpekar and Ozlem Kalinli and Parkin Kent and Parth Parekh and Paul Saab and Pavan Balaji and Pedro Rittner and Philip Bontrager and Pierre Roux and Piotr Dollar and Polina Zvyagina and Prashant Ratanchandani and Pritish Yuvraj and Qian Liang and Rachad Alao and Rachel Rodriguez and Rafi Ayub and Raghotham Murthy and Raghu Nayani and Rahul Mitra and Rangaprabhu Parthasarathy and Raymond Li and Rebekkah Hogan and Robin Battey and Rocky Wang and Russ Howes and Ruty Rinott and Sachin Mehta and Sachin Siby and Sai Jayesh Bondu and Samyak Datta and Sara Chugh and Sara Hunt and Sargun Dhillon and Sasha Sidorov and Satadru Pan and Saurabh Mahajan and Saurabh Verma and Seiji Yamamoto and Sharadh Ramaswamy and Shaun Lindsay and Shaun Lindsay and Sheng Feng and Shenghao Lin and Shengxin Cindy Zha and Shishir Patil and Shiva Shankar and Shuqiang Zhang and Shuqiang Zhang and Sinong Wang and Sneha Agarwal and Soji Sajuyigbe and Soumith Chintala and Stephanie Max and Stephen Chen and Steve Kehoe and Steve Satterfield and Sudarshan Govindaprasad and Sumit Gupta and Summer Deng and Sungmin Cho and Sunny Virk and Suraj Subramanian and Sy Choudhury and Sydney Goldman and Tal Remez and Tamar Glaser and Tamara Best and Thilo Koehler and Thomas Robinson and Tianhe Li and Tianjun Zhang and Tim Matthews and Timothy Chou and Tzook Shaked and Varun Vontimitta and Victoria Ajayi and Victoria Montanez and Vijai Mohan and Vinay Satish Kumar and Vishal Mangla and Vlad Ionescu and Vlad Poenaru and Vlad Tiberiu Mihailescu and Vladimir Ivanov and Wei Li and Wenchen Wang and Wenwen Jiang and Wes Bouaziz and Will Constable and Xiaocheng Tang and Xiaojian Wu and Xiaolan Wang and Xilun Wu and Xinbo Gao and Yaniv Kleinman and Yanjun Chen and Ye Hu and Ye Jia and Ye Qi and Yenda Li and Yilin Zhang and Ying Zhang and Yossi Adi and Youngjin Nam and Yu and Wang and Yu Zhao and Yuchen Hao and Yundi Qian and Yunlu Li and Yuzi He and Zach Rait and Zachary DeVito and Zef Rosnbrick and Zhaoduo Wen and Zhenyu Yang and Zhiwei Zhao and Zhiyu Ma},
      year={2024},
      eprint={2407.21783},
      archivePrefix={arXiv},
      primaryClass={cs.AI},
      url={https://arxiv.org/abs/2407.21783}, 
}

@misc{kirillov2023segment,
      title={Segment Anything}, 
      author={Alexander Kirillov and Eric Mintun and Nikhila Ravi and Hanzi Mao and Chloe Rolland and Laura Gustafson and Tete Xiao and Spencer Whitehead and Alexander C. Berg and Wan-Yen Lo and Piotr Dollár and Ross Girshick},
      year={2023},
      eprint={2304.02643},
      archivePrefix={arXiv},
      primaryClass={cs.CV},
      url={https://arxiv.org/abs/2304.02643}, 
}

@InProceedings{Sun_2024_CVPR,
    author    = {Sun, Zhonglin and Feng, Chen and Patras, Ioannis and Tzimiropoulos, Georgios},
    title     = {LAFS: Landmark-based Facial Self-supervised Learning for Face Recognition},
    booktitle = {Proceedings of the IEEE/CVF Conference on Computer Vision and Pattern Recognition (CVPR)},
    month     = {June},
    year      = {2024},
    pages     = {1639-1649}
}

@misc{devlin2019bertpretrainingdeepbidirectional,
      title={BERT: Pre-training of Deep Bidirectional Transformers for Language Understanding}, 
      author={Jacob Devlin and Ming-Wei Chang and Kenton Lee and Kristina Toutanova},
      year={2019},
      eprint={1810.04805},
      archivePrefix={arXiv},
      primaryClass={cs.CL},
      url={https://arxiv.org/abs/1810.04805}, 
}

@misc{yenduri2023generativepretrainedtransformercomprehensive,
      title={Generative Pre-trained Transformer: A Comprehensive Review on Enabling Technologies, Potential Applications, Emerging Challenges, and Future Directions}, 
      author={Gokul Yenduri and Ramalingam M and Chemmalar Selvi G and Supriya Y and Gautam Srivastava and Praveen Kumar Reddy Maddikunta and Deepti Raj G and Rutvij H Jhaveri and Prabadevi B and Weizheng Wang and Athanasios V. Vasilakos and Thippa Reddy Gadekallu},
      year={2023},
      eprint={2305.10435},
      archivePrefix={arXiv},
      primaryClass={cs.CL},
      url={https://arxiv.org/abs/2305.10435}, 
}

@inproceedings{NIPS2017_3f5ee243,
 author = {Vaswani, Ashish and Shazeer, Noam and Parmar, Niki and Uszkoreit, Jakob and Jones, Llion and Gomez, Aidan N and Kaiser, \L ukasz and Polosukhin, Illia},
 booktitle = {Advances in Neural Information Processing Systems},
 editor = {I. Guyon and U. Von Luxburg and S. Bengio and H. Wallach and R. Fergus and S. Vishwanathan and R. Garnett},
 pages = {},
 publisher = {Curran Associates, Inc.},
 title = {Attention is All you Need},
 url = {https://proceedings.neurips.cc/paper_files/paper/2017/file/3f5ee243547dee91fbd053c1c4a845aa-Paper.pdf},
 volume = {30},
 year = {2017}
}

@ARTICLE{6795963,
  author={Hochreiter, Sepp and Schmidhuber, Jürgen},
  journal={Neural Computation}, 
  title={Long Short-Term Memory}, 
  year={1997},
  volume={9},
  number={8},
  pages={1735-1780},
  doi={10.1162/neco.1997.9.8.1735}}

@inproceedings{10.1609/aaai.v38i8.28681,
author = {Fan, Jinyong and Shen, Yanyan},
title = {StockMixer: a simple yet strong MLP-based architecture for stock price forecasting},
year = {2024},
isbn = {978-1-57735-887-9},
publisher = {AAAI Press},
url = {https://doi.org/10.1609/aaai.v38i8.28681},
doi = {10.1609/aaai.v38i8.28681},
booktitle = {Proceedings of the Thirty-Eighth AAAI Conference on Artificial Intelligence and Thirty-Sixth Conference on Innovative Applications of Artificial Intelligence and Fourteenth Symposium on Educational Advances in Artificial Intelligence},
articleno = {933},
numpages = {9},
series = {AAAI'24/IAAI'24/EAAI'24}
}

@article{sheikh1996barra,
  title={BARRA’s risk models},
  author={Sheikh, Aamir},
  journal={Barra Research Insights},
  pages={1--24},
  year={1996}
}
\newpage
\section{Appendix}
\label{sec:appendix}
\begin{table}[ht]
\centering
\small
\begin{tabularx}{\columnwidth}{lX} 
\toprule
\textbf{Symbol} & \textbf{Description} \\
\midrule
$z_t$           & Continuous latent market regime state at time $t$. \\ \addlinespace[2pt]
$\mathcal{C}$   & Global set of financial and risk constraints. \\ \addlinespace[2pt]
$f$             & Formulaic alpha factor expressed as a symbolic DAG. \\ \addlinespace[2pt]
$\mathbf{w}_t$  & Target portfolio weights vector for $N$ assets. \\ \addlinespace[2pt]
$\mathbf{s}_t$  & Cross-sectional signal vector generated by alpha factor $f$. \\ \addlinespace[2pt]
$\theta$        & Trainable parameters of the domain-specific LLM (\textit{CQ2}). \\ \addlinespace[2pt]
$\lambda(z_t)$  & Regime-adaptive risk aversion coefficient (Numerical Channel). \\ \addlinespace[2pt]
$E$             & Evidence bundle (backtest results and execution diagnostics). \\ \addlinespace[2pt]
$\mathcal{M}$   & Market Dynamics Module (generates $z_t$). \\ \addlinespace[2pt]
$\mathcal{R}$   & Alpha Research Module (performs constrained MCTS). \\ \addlinespace[2pt]
$\mathcal{P}$   & Portfolio Construction Module (optimizes $\mathbf{w}_t$). \\ \addlinespace[2pt]
$\mathcal{E}$   & Execution Module (manages trade implementation). \\ \addlinespace[2pt]
$\mathcal{U}$   & Update Operator (manages the closed-loop feedback). \\ \addlinespace[2pt]
$G_{\text{forbidden}}$ & Subset of static syntax rules derived from $\mathcal{C}$. \\ \addlinespace[2pt]
$R_{\text{perf}}$ & Quantitative backtest performance metrics (e.g., Rank IC). \\ \addlinespace[2pt]
$R_{\text{model}}$ & Alignment score from the RLHF reward model. \\
\bottomrule
\end{tabularx}
\caption{Summary of key notations and module definitions in PandaAI.}
\label{tab:notations}
\end{table}

\subsection{Detailed Procedure of Constrainted LLM-guided Monte Carlo Tree Search}
\label{sec:cmcts}
Algorithm \ref{alg:llm_mcts} illustrates the pseudocode of the whole procedure.

\textbf{1. Selection (Regime-Adaptive):}
Nodes are selected using a modified UCT algorithm where the exploration constant $c$ is not static but modulated by the market state $z_t$:
\begin{equation}
    \text{UCT}(s) = \frac{Q(s)}{N(s)} + c(z_t) \cdot \sqrt{\frac{\ln N(\text{parent})}{N(s)}}
\end{equation}
Here, $c(z_t)$ is inversely proportional to the market entropy detected by $\mathcal{M}$. In stable regimes, $c(z_t)$ increases to encourage broad exploration; in turbulent regimes (e.g., liquidity crises), $c(z_t)$ decays to enforce conservative exploitation of known safe sub-trees.

\textbf{2. Expansion (Constrained Generation):}
The $\text{LLM}_\theta$ acts as the policy network $\pi(a|s, z_t)$. To operationalize \textbf{Constraints as A Priori Regularization}, we clarify the relationship between the static syntax rules $G_{\text{forbidden}}$ (in Algorithm 1) and the dynamic risk constraints $\mathcal{C}$ (updated by Module $\mathcal{U}$): specifically, $G_{\text{forbidden}} \subset \mathcal{C}$.
During generation, we employ a \textbf{"Prompt-Check-Regenerate"} loop:
\begin{itemize}
    \item \textbf{Prompt Injection:} The semantic rules from $\mathcal{C}$ are combined with the regime state $z_t$, which is \textbf{prepended as continuous soft tokens to the system prompt via Channel 1}.
    \item \textbf{Pre-Simulation Filter:} Generated candidates are immediately parsed against $G_{\text{forbidden}}$. Invalid formulas are rejected before entering the costly simulation phase.
\end{itemize}

\textbf{3. Simulation (Feedback \& Soft Penalty):}
Candidates passing the expansion filter undergo backtesting. Although obvious violations are filtered \textit{a priori}, subtle financial toxicities (e.g., high correlation with existing factors) can only be detected \textit{a posteriori}. Thus, we define the node value function $V(f)$ with a penalty term for these residual violations:
\begin{equation}
    V(f) = R_{\text{perf}}(f) + \alpha \cdot R_{\text{model}}(f) - \lambda \cdot \mathbb{I}(f \notin \mathcal{C}_{\text{dynamic}}) 
    \label{eq:simulation}
\end{equation}
where $R_{\text{perf}}$ denotes backtest metrics (e.g., IC), and $R_{\text{model}}$ is the alignment score from the RLHF Reward Model. The penalty $\mathbb{I}(\cdot)$ handles the dynamic subset $\mathcal{C}_{\text{dynamic}} = \mathcal{C} \setminus G_{\text{forbidden}}$, ensuring that factors surviving the filter but failing on risk metrics (e.g., excessive turnover $>50\%$ daily\footnote{High turnover rates usually incur prohibitive transaction costs.}) are heavily penalized.

\textbf{4. Backpropagation:}
The evaluation signals are propagated to update the node statistics, progressively steering the LLM towards the "valid and robust" subspace of the alpha universe.

\begin{figure*}[t]
\centering
\captionsetup{type=algorithm}
\caption{LLM-Guided Constrained MCTS for Alpha Mining}
\label{alg:llm_mcts}
\begin{minipage}{0.98\textwidth}
\begin{algorithmic}[1]
\Statex \textbf{Input:} $f_{\mathrm{seed}}$ (seed alpha), $\text{LLM}_\theta$, $z_t$ (market state), $\mathcal{C}$ (constraint set), $B$ (budget)
\Statex \textbf{Output:} $F_{\mathrm{zoo}}$ (robust alpha repository)

\Statex \textbf{/* Initialization */}
\State $F_{\mathrm{zoo}} \gets \emptyset$
\State $s_0 \gets \textsc{CreateRoot}(f_{\mathrm{seed}})$
\State $\textsc{Tree } \mathcal{T} \gets \{s_0\}$
\State $G_{\text{forbidden}} \gets \textsc{ExtractSyntaxRules}(\mathcal{C})$ 
\Comment{Static constraints subset}

\For{$\textit{iter} \gets 1$ \textbf{to} $B$}

\Statex \textbf{/* 1. Selection Phase (Regime-Adaptive via H1) */}
\State $c_{\text{exp}} \gets \textsc{ComputeExploration}(z_t)$ 
\Comment{$c(z_t)$ modulated by market entropy}
\State $s_{\text{leaf}} \gets \textsc{SelectViaUCT}(\mathcal{T}, s_0, c_{\text{exp}})$

\Statex \textbf{/* 2. Expansion Phase (Constraint-Guided via H2) */}
\State $\textit{context} \gets \textsc{GetContext}(s_{\text{leaf}})$
\State $\textit{prompt} \gets \textit{context} \cup \mathcal{C}$ 
\Comment{Inject constraints into prompt}
\State $f_{\text{new}} \gets \text{NULL}$

\Statex \textbf{/* Pre-Simulation Filter Loop */}
\While{$f_{\text{new}}$ is Invalid OR $f_{\text{new}} \in G_{\text{forbidden}}$}
    \State $f_{\text{new}} \gets \text{LLM}_\theta.\textsc{Generate}(\textit{prompt}, z_t)$ 
    \Comment{CoT generation}
\EndWhile
\State $s_{\text{new}} \gets \textsc{AddChild}(\mathcal{T}, s_{\text{leaf}}, f_{\text{new}})$

\Statex \textbf{/* 3. Simulation Phase (Value Estimation) */}
\State $R_{\text{perf}} \gets \textsc{Backtest}(f_{\text{new}}, z_t)$
\State $R_{\text{model}} \gets \textsc{RewardModel}(f_{\text{new}})$ 
\Comment{RLHF alignment score}
\State $\textit{penalty} \gets \textsc{CheckDynamicConstraints}(f_{\text{new}}, \mathcal{C} \setminus G_{\text{forbidden}})$
\State $V(f_{\text{new}}) \gets R_{\text{perf}} + \alpha \cdot R_{\text{model}} - \lambda \cdot \textit{penalty}$ 
\Comment{Eq. 8}

\Statex \textbf{/* 4. Backpropagation Phase */}
\State $\textsc{BackupValues}(\mathcal{T}, s_{\text{new}}, V(f_{\text{new}}))$

\Statex \textbf{/* Repository Update */}
\If{$V(f_{\text{new}}) > \tau_{\text{accept}}$}
    \State $F_{\mathrm{zoo}} \gets F_{\mathrm{zoo}} \cup \{f_{\text{new}}\}$
\EndIf

\EndFor
\State \Return $F_{\mathrm{zoo}}$

\end{algorithmic}
\end{minipage}
\end{figure*}
\subsection{Detailed Fine-Tuning $\mathcal{T}$}
\label{sec:apendix_finetune}
The fine-tuning dataset is not publicly available due to privacy obligations to clients and restrictions imposed by non-disclosure agreements.
\subsubsection{Supervised Fine-Tuning}
\paragraph{Regime-Conditioned Instruction Tuning} A financial instruction dataset $\mathcal{D}_{\text{SFT}}$ is constructed where each sample $v$ is tagged with the originating market state $z_t$ from Module $\mathcal{M}$. Each sample $v$ contains three components $(x,y, z_t)$, where $x$ denotes the question and $y$ corresponds to the answer. This teaches the model to condition financial concepts on market context. The same underlying concept must be implemented differently depending on $z_t$, supporting \textbf{H1} by embedding regime-awareness into the model's generative process.

\paragraph{Chain-of-Thought Financial Reasoning} A reasoning trace dataset $V_{SFT\_QA}$ is constructed, where each sample is a triplet $v=(x,c,y)$. $c$ represents the reasoning trace formatted as $<\textit{think}>\dots</\textit{think}>$ and $y$ corresponds to the answer formatted as $<\textit{answer}>\dots</\textit{answer}>$. Through $\mathcal{D}_{\text{QRA}}$ samples, explicit reasoning structures are learned which decompose factor design into logical steps. 


\paragraph{Supervised Fine-Tuning}
Supervised Fine-Tuning (SFT) is initially performed on DeepSeek-Coder-33B using $V_{SFT\_QA}$ to broaden LLM's perspective on the contextualization hypothesis. In this stage, the tagged $z_t$ of each sample is injected into LLM, following \textbf{Channel 1}, and parameters of the Symbolic Adapter are trained synchronously with \textit{CQ2} fine-tune. Next, SFT continues using $V_{SFT\_QRA}$, optimizing key aspects of financial reasoning. However, naively implementing SFT on all the parameters with new data leads to the catastrophic forgetting \cite{li-etal-2024-revisiting} that hampers the performance of the original models. To solve this limitation, the original DeepSeek-Coder-33B is used as the teacher, and an auxiliary loss $L_{KD}$ is used to measure the KL distance between the output of the original and fine-tuned models. 

\subsubsection{Execution-Driven Reinforcement Learning from Human Feedback}
While SFT establishes financial knowledge, Reinforcement Learning from Human Feedback aligns \textit{CQ2}'s generative behavior with execution success criteria derived from trading contexts. Conventional RLHF is extended by incorporating execution simulation feedback into the reward signal.

\paragraph{RLHF Dataset}
Two different datasets are produced and used in the RLHF procedure: (1) Reward Model Dataset $V_{RL\_RM}$, where one sample consists of a prompt $x$ and two user responses $(y_0,y_1)$, used to train our reward model, and (2) Proximal Policy
Optimization Dataset $V_{RL\_PPO}$, with only prompt $x$ of each sample, which are used as inputs for RLHF.

\paragraph{Execution-Grounded Reward Modeling (RM)}
DeepSeek-R1-Distill-Qwen-7B \cite{deepseekai2025deepseekr1incentivizingreasoningcapability} is fine-tuned, followed by a randomly initialized linear head that outputs a scalar value, to take in a prompt and response. In this work, only a 7B reward model is used to reduce computational cost, and larger reward models can be unstable and are therefore less suitable as the value function during reinforcement learning. \cite{10.5555/3600270.3602281}. This model is trained to predict which $y \in \{y_0, y_1\}$ is better as judged by a user, given a prompt $x$. The RM is trained to compare two model outputs
on the same prompt. If the response preferred by the user is $y_j$, the RM loss can be written as:
\begin{equation}
    L(r_\theta) = - \mathbb{E}_{V_{RM}}\left[\log(\sigma(r_\theta(x, y_i) - r_\theta(x, y_{1-i})) \right],
\end{equation}
where $r_{\theta}(x,y)$ is the scalar output of the reward model for the prompt $x$ and the response $y$ with parameters $\theta$, and $\sigma(z) = \frac{1}{1+e^{-z}}$ is the logistic sigmoid function. This RM has been proven to generalize well on unseen datasets \cite{10.5555/3495724.3495977}. Therefore, this RM can guarantee the quality of the following Reinforcement Learning. Furthermore, $r_{\theta}(\cdot, \cdot)$ is also used as $R_{model}(f)=r_{\theta}(x,f)$ in Equation (\ref{eq:simulation}) of Section \ref{subsec:alpha_research}, where $x$ is the prompt and $f$ is the generated formula factor. Maximizing $R_{model}(f)$ can make the generated factor satisfy the user preference.

\paragraph{Reinforcement Learning (RL)}
After training the reward model, a policy $\pi^{RL}_\phi(y \mid x)$ is optimized to generate responses that achieve higher reward under human preference. This stage is formulated as reinforcement learning in a bandit-style setting: a prompt $x$ is sampled from the dataset, the policy generates a complete response $y \sim \pi^{RL}_\phi(\cdot \mid x)$, and the episode terminates with a single scalar score provided by the fixed reward model $r_\theta(x,y)$.

Directly maximizing $r_\theta(x,y)$ can cause the policy to drift away from the SFT policy and overfit to imperfections in the reward model. To stabilize training, a KL regularization term that penalizes deviation from the SFT policy $\pi^{SFT}(y \mid x)$ is introduced. The resulting shaped reward is
\begin{equation}
R(x,y)= r_\theta(x,y)\;-\;\beta \log\!\left(\frac{\pi^{RL}_\phi(y\mid x)}{\pi^{SFT}(y \mid x)}\right),
\end{equation}
where $\beta$ controls the strength of the constraint. This KL term acts as an entropy bonus, encouraging the policy to explore and deterring it from collapsing to a single mode. Then $\mathbb{E}_{y\sim \pi_\phi(\cdot\mid x)}[R(x,y)]$ is maximized using Proximal Policy Optimization (PPO) \cite{schulman2017proximalpolicyoptimizationalgorithms}.
\begin{table*}[t]
\small
\centering
\begin{tabular}{
  >{\centering\arraybackslash}m{3.5cm}
  >{\raggedright\arraybackslash}p{11cm}
}
\toprule
\textbf{Factor Name} & \textbf{Factor Formula} \\
\midrule
\textbf{Factor 1} & {((-1 $\times$ RANK(DELTA(RETURNS, 3))) $\times$  CORRELATION(OPEN, VOLUME, 10))} \\
\midrule
\textbf{Factor 2}  & {INDUSTRY\_NEUTRALIZE(((CORRELATION(DELTA(CLOSE, 1), DELTA(DELAY(CLOSE, 1), 1), 250) * DELTA(CLOSE, 1)) / CLOSE)) / SUM(((DELTA(CLOSE, 1) / DELAY(CLOSE, 1))**2), 250))} \\
\midrule
\textbf{Factor 3}  & {((-1 * RANK(TS\_RANK(CLOSE, 10))) * RANK(DELTA(DELTA(CLOSE, 1), 1))) * RANK(TS\_RANK((VOLUME / ADV(20)), 5)))} \\
\midrule
\textbf{Factor 4}  & {(-1 * SIGN(((CLOSE - DELAY(CLOSE, 7)) + DELTA(CLOSE, 7)))) * (1 + RANK((1 + SUM(RETURNS, 250))))} \\
\midrule
\textbf{Factor 5}  & {(0 - (1 * ((2 * SCALE(RANK(((((CLOSE - LOW) - (HIGH - CLOSE)) / (HIGH - LOW)) * VOLUME)))) - SCALE(RANK(TS\_ARGMAX(CLOSE, 10))))))} \\
\midrule
\textbf{Factor 6}  & {(-1 * SUM(RANK(CORRELATION(RANK(HIGH), RANK(VOLUME), 3)), 3))} \\
\midrule
\textbf{Factor 7}  & {IF((0 < TS\_MIN(DELTA(CLOSE, 1), 5)), DELTA(CLOSE, 1), (IF((TS\_MAX(DELTA(CLOSE, 1), 5) < 0), DELTA(CLOSE, 1), (-1 * DELTA(CLOSE, 1)))))} \\
\midrule
\textbf{Factor 8}  & {((-1 * RANK((OPEN - DELAY(HIGH, 1)))) * RANK((OPEN - DELAY(CLOSE, 1)))) * RANK((OPEN - DELAY(LOW, 1)))} \\
\bottomrule
\end{tabular}
\caption{The summary of Formulaic Alpha Factors in our experiments}
\label{tab:factors}
\end{table*}
\subsection{Evaluation Metrics}
\label{sec:eval metrics}
To comprehensively evaluate the performance of the proposed Panda AI framework in quantitative finance decision-making, we employ three groups of widely recognized metrics, measuring its predictive power, risk-adjusted performance, and absolute performance, respectively.

\paragraph{Predictive Power}
Predictive power metrics quantify the correlation between the alpha signals generated by the model and the subsequent realized returns. We employ the Information Coefficient and its variant.
\begin{itemize}
    \item \textbf{Information Coefficient (IC)} measures the linear correlation between the predicted returns vector $R_p$ and the actual realized returns vector $R_a$, defined as the Pearson correlation coefficient:
    \begin{equation}
        IC = \rho(R_p, R_a)
    \end{equation}
    where $\rho(\cdot)$ denotes the Pearson correlation. An IC value closer to 1 indicates stronger predictive ability.

    \item \textbf{Rank Information Coefficient (Rank IC)} assesses the monotonic relationship by comparing the ordinal rankings of predictions and outcomes, which is more robust to outliers:
    \begin{equation}
        RankIC = \rho(\text{rank}(R_p), \text{rank}(R_a))
    \end{equation}
    where $\text{rank}(\cdot)$ denotes the transformation of the return series into their rank orders.
\end{itemize}

\paragraph{Risk-Adjusted Performance}
Risk-adjusted performance metrics evaluate the model's ability to generate excess return per unit of risk undertaken and the stability of its predictive skill.
\begin{itemize}

    \item \textbf{Information Coefficient Information Ratio (ICIR)} evaluates the stability and significance of predictive skill over time. It is defined as the mean of period IC divided by its standard deviation:
    \begin{equation}
        ICIR = \frac{\overline{IC}}{\sigma(IC)}
    \end{equation}
    where $\overline{IC}$ is the mean of IC over multiple periods (e.g., monthly), and $\sigma(IC)$ is its standard deviation. A higher ICIR indicates more stable and reliable predictive ability.
\end{itemize}

\paragraph{Absolute Performance}
Absolute performance metrics provide direct insight into the raw profitability and the extreme downside risk of the investment strategy.
\begin{itemize}
    \item \textbf{Annualized Return (AR)} compounds the cumulative return to an annualized figure, facilitating comparison across strategies and time periods:
    \begin{equation}
        AR = (1 + \text{Total Return})^{\frac{252}{N}} - 1
    \end{equation}
    where \textit{Total Return} is the cumulative return over the entire period, and $N$ is the total number of trading days. The constant 252 represents the typical number of trading days in a year.

    \item \textbf{Maximum Drawdown (MDD)} quantifies the maximum observed loss from a peak to a trough of a portfolio before a new peak is attained. It is a critical measure of extreme downside risk:
    \begin{equation}
        MDD = \max_{t \in (0,T)} \left( \frac{P_t - \min_{t \leq \tau \leq T}(P_\tau)}{P_t} \right)
    \end{equation}
    where $P_t$ is the portfolio's net asset value at time $t$, and $T$ is the evaluation period.
\end{itemize}

\end{document}